\def\year{2020}\relax
\documentclass[letterpaper]{article} 
\usepackage{aaai20}  
\usepackage{times}  
\usepackage{helvet} 
\usepackage{courier}  
\usepackage[hyphens]{url}  
\usepackage{graphicx} 
\usepackage{graphicx}  
\usepackage{amsmath,amsfonts}
\usepackage{booktabs, multicol, multirow}
\usepackage{subcaption}

\newcommand{\citet}[1]{\citeauthor{#1}~\shortcite{#1}}
\newcommand{\citep}{\cite}

\usepackage{titlesec}
\titlespacing{\paragraph}{%
  0pt}{
  0.5em}{
  1em}

\nocopyright
\title{Combating False Negatives in Adversarial Imitation Learning}

\author{Konrad \.Zo\l{}na\textsuperscript{\rm 1, 2},
Chitwan Saharia\textsuperscript{\rm 2, 4},
L\'eonard Boussioux\textsuperscript{\rm 2, 5, 6},\\
{\Large \bf
David Yu-Tung Hui\textsuperscript{\rm 2},
Maxime Chevalier-Boisvert\textsuperscript{\rm 2},
Dzmitry Bahdanau\textsuperscript{\rm 2, 3, 6},
Yoshua Bengio\textsuperscript{\rm 2, 3, 8}}\\\\
\textsuperscript{\rm 1}{Jagiellonian University}, \textsuperscript{\rm 2}{Mila},
\textsuperscript{\rm 3}{U. Montreal},
\textsuperscript{\rm 4}{IIT Bombay}, \textsuperscript{\rm 5}{MIT}, \\ \textsuperscript{\rm 6}{\'Ecole CentraleSup\'elec},
\textsuperscript{\rm 7}{Element AI}, \textsuperscript{\rm 8}{CIFAR Senior Fellow}\\\\
konrad.zolna@gmail.com
}

\begin{document}

\maketitle

\begin{abstract}
In adversarial imitation learning, a discriminator is trained to differentiate agent episodes from expert demonstrations representing the desired behavior. However, as the trained policy learns to be more successful, the negative examples (the ones produced by the agent) become increasingly similar to expert ones. Despite the fact that the task is successfully accomplished in some of the agent's trajectories, the discriminator is trained to output low values for them. We hypothesize that this inconsistent training signal for the discriminator can impede its learning, and consequently leads to worse overall performance of the agent. We show experimental evidence for this hypothesis and  that the 'False Negatives' (i.e. successful agent episodes) significantly hinder adversarial imitation learning, which is the first contribution of this paper. Then, we propose a method to alleviate the impact of false negatives and test it on the BabyAI environment. This method consistently improves sample efficiency over the baselines by at least an order of magnitude.
\end{abstract}

\section{Introduction}

Progress in Deep Reinforcement Learning is impeded by the necessity of handcrafting reward functions, which may be especially difficult for grounded language tasks~\citep{overview_language}. To avoid using a reward function to judge the agent's behavior, Imitation Learning (IL) trains an agent to mimic an expert's policy using demonstrations. The simplest version of IL, Behavioral Cloning (BC) \citep{pomerleau89alvinn}, trains a policy to regress expert actions from demonstrations in a supervised setup. This approach is appealing due to its simplicity but suffers from the problem of compounding errors \citep{ross2011reduction}.

Another IL method, Generative Adversarial Imitation Learning (GAIL) ~\citep{ermon_gail}, jointly learns reward functions and training policies. GAIL trains a discriminator to differentiate agent from expert trajectories, which simultaneously acts as a reward function. Hence, the agent tries to act more and more like the expert in order to fool the discriminator and get a higher reward.

While GAIL works well in the initial phase of the learning procedure, as the agent reaches high success rates on the given task, we observed that its performance tends to be unstable. We hypothesize that this is due to the fact that the discriminator has to classify successful agent episodes as negative examples, even though they are very similar to expert demonstrations. This problem is negligible during the initial phase when the agent executes a random policy, but as the policy improves, the number of such successful trajectories labeled as non-expert increases. We refer to this phenomenon as the \emph{False Negatives}~(FN) problem, as successful agent trajectories are falsely labeled as negative examples.

\begin{figure}[t!]
    \centering
    \begin{subfigure}[t]{0.3\linewidth}
        \centering
        \includegraphics[width=0.95\linewidth]{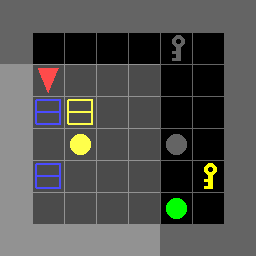}
        \caption{GoToLocal\\\textit{go to the yellow key}}
    \end{subfigure}%
    ~ 
    \begin{subfigure}[t]{0.3\linewidth}
        \centering
        \includegraphics[width=0.95\linewidth]{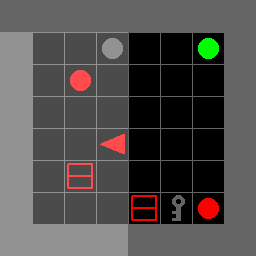}
        \caption{PickupLoc\\\textit{pick up a box\\in front of you}}
    \end{subfigure}%
    ~ 
    \begin{subfigure}[t]{0.3\linewidth}
        \centering
        \includegraphics[width=0.95\linewidth]{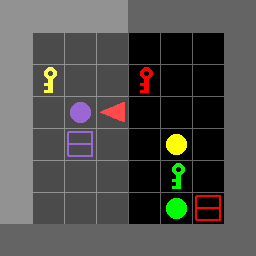}
        \caption{PutNextLocal\\\textit{put the green ball\\next to the red key}}
    \end{subfigure}
    \vspace{-0.15cm}
    \caption{Instructions and initial states for three BabyAI tasks. In (b), the reward for the episode can not be viewed as a function of the last observation, and hence memory is required for the discriminator in GAIL. Best viewed in color.
    \label{fig:env}}
    \vspace{-0.4cm}
\end{figure}

Our first contribution is a diagnostic method which shows that the FN problem significantly hinders adversarial imitation learning and measures the effect's strength. We use the BabyAI platform~\citep{babyai_iclr19} as the testbed since it is well-suited to analyse imitation learning sample efficiency. We focus on four levels of increasing difficulty (see Figure~\ref{fig:env}) and show that a naive application of GAIL is not able to solve any level due to false negatives.

We furthermore propose a method that can be applied for any goal-conditioned task and, by leveraging its multi-goal nature, addresses the FN problem. In particular, we simultaneously train two discriminators which take roles of reward functions as in ordinary GAIL. We use goal-conditioning to ensure that the FN problem does not occur when the first discriminator is trained. The second discriminator is trained to penalize the agent for exploiting inaccuracies of the first one. We show that the proposed technique enables the agent's performance to approach a $100\%$ success rate, while a naive application of GAIL fails.

\section{Background}\label{sec:background}

We study the problem of training agents to follow natural language instructions grounded in a partially-observable environment using adversarial imitation learning. In particular, we assume that $N$ pairs $(c_i, \tau_i)$ of instructions $c_i$ and their respective trajectories $\tau_i=\left( (o_1^i, a_1^i) , \, \ldots (o_{T(i)}^i, a_{T(i)}^i) \right)$ are provided for the agent to learn from. Here, $o_t^i$ and $a_t^i$ are the observation and action at the time step $t$, and $T(i)$ is the $i$-th episode length.

GAIL \citep{ermon_gail} is an imitation learning framework which combines ideas from Inverse Reinforcement Learning \citep{ng,abbeel} and Generative Adversarial Networks (GANs) \citep{gan}. In GAIL, the actor network represents the agent's policy $\pi$ while the discriminator network $D$ serves as a local reward function differentiating between the expert policy $\pi_E$ and the actor using observation-action pairs $(o, a)$. \citet{ermon_gail} define the optimization objective of GAIL as follows:
\begin{align}\label{eq:gail}
    \max _ { \pi } \min _ { D } & \Big( \mathbb { E } _ { \pi } [ -\log (1 - D ( o , a )) ] + \\ \nonumber
    & + \mathbb { E } _ { \pi _ { E } } [ -\log D ( o , a ) ] + \lambda H ( \pi ) \Big),
\end{align}
where function $H$ stands for the maximum entropy introduced by \citet{ziebart_entropy}, ensuring adequate exploration of the environment. GAIL was originally proposed for fully observable environments where the observation $o$ in Equation~\ref{eq:gail} must capture the complete state of the environment.

Similarly to \citet{fu_language_2019}, we extend GAIL to learn instruction-conditioned policies by conditioning both the generator (policy) $\pi$ and the discriminator $D$ on the instruction $c$.
The resulting objective is
\begin{align}
    \label{eq:orig_gail}
    \max _ { \pi } \min _ { D } & \Big( \mathbb { E } _ { \pi} [ -\log (1 - D ( o, c, a )) ] + \\ \nonumber
    & + \mathbb { E } _ { \pi _ { E }} [ -\log D ( o, c, a ) ] + \lambda H ( \pi ) \Big),
\end{align}
where $\mathbb { E } _ { \pi }$ stands for sampling $(o, c, a)$ by acting with the policy $\pi(a|o,c)$ conditioned on both the observation $o$ and the instruction $c$.

\subsection{Conditioned Recurrent Discriminator}\label{subsec:conditioned}

For partially-observable environments it is necessary to equip the discriminator with a memory to model the true environment reward (see Figure~\ref{fig:env}(b) for example). To this end, we implement the discriminator as a recurrent neural network. Its input is not a single observation-action pair $(o_i, a_i)$, as in the original GAIL formulation, but a full trajectory $\tau =  \left( (o_1, a_1) , \ldots , (o_T, a_T) \right)$. To address the goal-conditioned nature of the problem, the discriminator is still conditioned on the instruction $c$, as described in the previous section. Since our recurrent discriminator receives only full trajectories as the training signal, its predictions for incomplete sub-trajectories are not reliable, and hence its outputs should not be used to reward the policy at each intermediate step like Equation \ref{eq:orig_gail} would require.
For this reason, we reward the policy, which is also implemented as a recurrent network, only at the last step of the episode. The resulting training objective is
\begin{align}
    \label{eq:full_traj_gail}
    \max _ { \pi } \min _ { D } & \Big( \mathbb { E } _ { \pi} [ -\log (1 - D (c, \tau)) ] + \\ \nonumber
    & + \mathbb { E } _ { \pi _ { E }} [ -\log D (c, \tau) ] + \lambda H ( \pi ) \Big).
\end{align}

An alternative approach to adapt GAIL to partially observable environments would be to train the discriminator to distinguish not just between complete trajectories but also between incomplete sub-trajectories. The policy could then be rewarded at each step like it is done in the original GAIL.
Such a dense reward formulation may be necessary when the discriminator does not have memory, but with memory in place the dense rewards become optional. We have chosen the full trajectory approach to avoid some of the known issues of the dense reward GAIL formulation, such as the survival bias \citep{kostrikov2018discriminator}. In Section~\ref{subsec:exp-sub-trajectories}, we present additional experiments in which incomplete sub-trajectories are used to train the discriminator

To compute a batched version of the discriminator loss we use two finite buffers $B_{agent}$ and $B_{expert}$ that contain  agent trajectories and expert demonstrations respectively: 
\begin{align}\label{eq:memory}
    L_D(\theta) = & \mathbb{E}_{(c,\tau) \sim B_{agent}} - \log(1 - D_\theta(c,\tau)) \\ \nonumber
                + & \mathbb{E}_{(c,\tau) \sim B_{expert}} - \log(D_\theta(c,\tau)).
\end{align}

\section{False Negatives}\label{sec:false_negatives}

The main premise of GAIL is that discriminator-based rewards will be high for the episodes that manifest expected behaviour. In practice, however, the discriminator is trained to output higher rewards for the expert and lower for the agent which, unfortunately, is not the same.

In the starting phase of the learning procedure, the discriminator can reliably assign a high value to a successful episode. Indeed, such episode must be an expert demonstration as the agent always fails. However, the streams of positive and negative examples that the discriminator receives can become very similar as the agent gets better and more likely to succeed. The discriminator can no longer assume that all successful trajectories are coming from the expert and has to detect idiosyncratic features in expert demonstrations that are not necessarily related to solving a given task. In the extreme case of the perfect agent, the discriminator \emph{has} to overfit the nuisances in expert demonstrations to minimize its loss, because the agent and expert behaviours are indistinguishable. Discriminator-based rewards for successful agent episodes can potentially become as low as for unsuccessful ones.

The culprit of the aforementioned problem is a naive labeling of successful agent episodes as negative examples for the discriminator. We would call such examples \mbox{\textit{False Negatives} (FN)}.

\subsection{Oracle Filtering}

We propose an approach, which we call \textit{Oracle Filtering} (OF), to diagnose if the false negatives hinder the performance of adversarial imitation learning. In particular, OF uses the environment's true reward signal to identify the successful trajectories generated by the agent and filter them out. In this setting, the buffer of negative examples $B_{agent}$ contains only unsuccessful trajectories. The discriminator's loss is given as in Equation~\ref{eq:memory} but instead of $B_{agent}$ we use $B^{oracle}_{agent}$ which is a subset of $B_{agent}$ that contains unsuccessful trajectories only:
\begin{align}
    L^O_D(\theta) = & \mathbb{E}_{(c,\tau) \sim B^{oracle}_{agent}} - \log(1 - D_\theta(c,\tau)) \\ \nonumber
                + & \mathbb{E}_{(c,\tau) \sim B_{expert}} - \log(D_\theta(c,\tau)).
\end{align}
Note that when the OF method is applied, the environment rewards are used solely to filter out successful agent trajectories -- the agent does \emph{not} have access to them.

We found that using OF significantly improves the performance, and reduces sample complexity by at least an order of magnitude. This confirms our hypothesis that false negatives have a detrimental effect on GAIL training.

\subsection{Fake Conditioning}

Since the Oracle Filtering technique requires access to environment rewards, it is only a diagnostic method suited to assess the impact of false negatives. 
We propose a technique that does not need environment rewards to address the FN problem. Our technique is aimed at tasks where the policy is goal-conditioned. In our case, we assume that the policy is conditioned on language instructions but the technique is general and can be applied in any multi-goal setup.

First, we maintain a set of possible language instructions $S$ initialized from all the unique instructions in the expert demonstrations and, optionally, updated with the instructions collected when the agent interacts with the environment. Secondly, for each instruction-trajectory pair $(c,\tau)$, we replace its original instruction $c$ with $\tilde{c}$ that is randomly sampled from $S \backslash \{c\}$.

We call this technique \textit{Fake Conditioning} (FC). It is motivated by the fact that the success of a trajectory is conditioned on the instruction. Therefore, for each trajectory, by replacing the instruction with a new one, we produce a new instruction-trajectory pair that is very likely to be unsuccessful, even when it is successfully conditioned on the original instruction. The FC technique can be used to prepare discriminator training data with greatly limited number of the false negatives.

\paragraph{Expert FC} We call \textit{Expert FC} a technique whereby the discriminator is trained to distinguish the expert trajectories with the original instructions from fake conditioned expert trajectories. In this case, the agent trajectories are not used at all, and hence the training is no longer adversarial. The discriminator's loss is given as follows: 
\begin{align}
    L^E_D(\theta) = & \mathbb{E}_{(c,\tau) \sim B_{expert},\,\tilde{c} \sim S \backslash \{c\}} - \log(1 - D_\theta(\tilde{c},\tau)) \\ \nonumber
                + & \mathbb{E}_{(c,\tau) \sim B_{expert}} - \log(D_\theta(c,\tau)),
\end{align}
where $(\tilde{c}, \tau)$ is fake conditioned instruction-trajectory pair.

\paragraph{Agent FC} The FC technique can also be used with agent trajectories. In this case, which we call \textit{Agent FC}, all agent trajectories are fake conditioned while expert ones are left unaltered. It means that positive examples stay as in Expert FC formulation but negatives are constructed from the fake conditioned agent trajectories instead of expert ones. In this case, the discriminator's loss is given as follows:
\begin{align}
    L^A_D(\theta) = & \mathbb{E}_{(c,\tau) \sim B_{agent}, \,\tilde{c} \sim S \backslash \{c\}} - \log(1 - D_\theta(\tilde{c},\tau)) \\ \nonumber
                + & \mathbb{E}_{(c,\tau) \sim B_{expert}} - \log(D_\theta(c,\tau)).
\end{align}

We note that the policy that generates agent trajectories is always conditioned on the original instruction, for both Expert FC and Agent FC. The fake instruction-trajectory pairs are used to train the discriminator only.

\begin{table*}[ht]
\centering
\caption{Comparison of positives and negatives used for different type of discriminators.}
\smallskip
\small
\begin{tabular}{ll||c|c|c|c|c}
 & & {\bf Conventional} & {\bf Agent FC} & {\bf Expert FC} & {\bf Done Detector} & {\bf Blank Conditioning} \\
 \midrule \midrule
 \multirow{2}{*}{\bf Positives} & trajectories & \multicolumn{4}{c|}{expert} & expert \\
 & {\it instructions} & \multicolumn{4}{c|}{{\it original}} & {\it blank} \\
 \midrule \midrule
 \multirow{2}{*}{\bf Negatives} & trajectories & agent & agent & expert & incomplete expert & agent \\
 & {\it instructions} & {\it original} & {\it fake} & {\it fake} & {\it original} & {\it blank} \\
\end{tabular}
\label{tab:diagram}
\end{table*}

\subsection{Auxiliary Rewards}

The FC technique greatly reduces the percentage of FN. Unfortunately, the discriminator is no longer trained on agent trajectories with the original instructions, and hence the rewards that the discriminator issues to the policy can potentially be less accurate. These reward inaccuracies could be exploited by the agent.

Another potential problem can occur when the fakeness of an instruction-trajectory pair can be inferred solely from the initial observation (without the agent's action). This could be used by the discriminator to surely identify negative examples since only these are false conditioned. For example, when the instruction requires reaching an object that is not present in the scene, the discriminator could spot that and reward the agent lowly regardless of its actions.

To address the aforementioned problems, we rely on auxiliary rewards provided by additional discriminators. We design these trainable auxiliary rewards to discourage degenerate behaviours that may arise from using the FC technique.

In this section we propose two possible ways of training an additional discriminator, which are \emph{Blank Conditioning} and \emph{Done Detector}. Together with previously proposed Agent FC and Expert FC techniques, four different discriminators can be trained. Each discriminator is trained separately to minimize its own loss and provides rewards of the form $- \log(1 - D_\theta(\cdot))$. In practice, a small $\epsilon>0$ is added to the discriminator's predictions to make the rewards bounded.

\paragraph{Blank Conditioning} When Expert FC is used, the discriminator's training distribution includes only expert trajectories. As result, its predictions on agent-generated trajectories may be less accurate which can be exploited by the policy. To address this problem, we propose to train an extra discriminator which distinguishes between agent and expert trajectories but is \emph{not} conditioned. It means that the agent is additionally rewarded for generating trajectories which resemble the expert (for \emph{a} goal/instruction, \emph{not} the given goal/instruction). To get the highest rewards, its trajectories have to match the distribution of expert trajectories on which the main discriminator is trained. The auxiliary reward for Blank Conditioning is based on the discriminator trained with the following loss:
\begin{align}
    L^B_D(\theta) = & \mathbb{E}_{(c,\tau) \sim B_{agent}} - \log(1 - D_\theta(c_\emptyset,\tau)) \\ \nonumber
                + & \mathbb{E}_{(c,\tau) \sim B_{expert}} - \log(D_\theta(c_\emptyset,\tau)),
\end{align}
where $c_\emptyset$ is a fixed blank instruction that masks the original ones\footnote{In theory the conditioning on the instruction could be skipped but using $c_\emptyset$ makes sharing the architecture straightforward -- the extra reward model can be implemented as the extra discriminator head.}. We call this technique \textit{Blank Conditioning}.

Unfortunately, Blank Conditioning can be impeded by the FN problem as well. However, we hypothesize that the repercussions will be less severe because only an auxiliary reward is affected. In particular, even when the agent masters the task and generates a lot of FN in the auxiliary discriminator's training data, the main reward will still be informative since the main discriminator uses training data without FN.

\paragraph{Done Detector} The second auxiliary discriminator is trained to detect if a given trajectory is finished. Its negative examples are unfinished expert demonstrations (with the original instruction), while the finished ones are positive examples. In both positive and negative examples the original instruction is used. Hence, the loss forces the discriminator to focus on both the trajectory and the instruction to understand if the goal is reached. The loss is the following:
\begin{align}
    L^D_D(\theta) = & \mathbb{E}_{(c,\tau) \sim B^{sub}_{expert}} - \log(1 - D_\theta(c,\tau)) \\ \nonumber
                + & \mathbb{E}_{(c,\tau) \sim B_{expert}} - \log(D_\theta(c,\tau)),
\end{align}
where $B^{sub}_{expert}$ is built out of $B_{expert}$ and contains all possible incomplete sub-trajectories only (i.e. each trajectory from $B^{sub}_{expert}$ comes from the element of $B_{expert}$ but is cut before they have finished). We call this discriminator \textit{Done Detector}.

We note that the loss for Done Detector does not depend on the agent's performance, and hence the extra discriminator is not trained adversarially. There are also no false negatives since all negative examples are constructed in a controlled way and are never successful.

Blank Conditioning is supposed to address the main limitation of Expert FC (lack of agent trajectories in discriminator training data), while Done Detector is for Agent FC (fake instructions with agent trajectories may be too easy to identify). However, both auxiliary rewards can be used with any FC method. It means that we can train all four discriminators simultaneously (one for each loss introduced in this subsection) to get four different rewards which can be mixed together to train the agent.

In Table~\ref{tab:diagram}, we provide the overview of positives and negatives used to train discriminators presented in this work. OF is skipped as it is only a diagnostic tool.

\section{Related Work}

\paragraph{Instruction Following with Natural Language} We consider the problem of following natural language instructions. Recently, reinforcement learning methods have been applied to make progress in this area across a variety of environments~\citep{chaplot,hermann,street_bc}. In these approaches, an agent is rewarded when it successfully follows an instruction. However, designing the respective reward function is non-trivial~\citep{overview_language}. Formalizing the completion semantics, implicit in natural language instruction, has been an open research problem since early efforts~\citep{winograd}. Human handcrafting of the reward function becomes increasingly infeasible as the number of instructions and complexity of the environment scale. Applying IL methods in this area is therefore promising~\citep{duvallet2013imitation}.

\paragraph{Imitation Learning} As mentioned before, IL alleviates the need to handcraft reward functions by learning policies from demonstrations. In this work we consider Behavioral Cloning (BC,~\citep{pomerleau89alvinn}) as a baseline and our method improves Generative Adversarial Imitation Learning (GAIL) \citep{ermon_gail} which is closely related to Inverse Reinforcement Learning~\cite{abbeel2004apprenticeship}. While all imitation learning methods can be used for learning instruction-conditioned policies \citep{mei_listen_2016,fu_language_2019,bahdanau_learning_2019}, prior literature only features instructions that can be verified by the final observation and does not discuss the topic of converging to near-perfect performance (with the exception of \citep{bahdanau_learning_2019}, see more on that below).

\paragraph{False Negatives} We found that the False Negative problem significantly impedes adversarial imitation learning. As such, the issue was also found problematic in prior work \citep{zolna2019taskrelevant}. However, prior work lacked a diagnostic tool to assess the impact of the FN problem and estimate how much performance can be improved if the problem is addressed. \citet{zolna2019taskrelevant} proposed a heuristic method to address the FN problem called Actor Early Stopping. This method terminates well performing episodes to limit the number of successful agent states used to train the discriminator. AGILE proposed by \citet{bahdanau_learning_2019} is another heuristic solution which deals with False Negatives in the very similar setup of training a reward model. The method filters out states that were most highly rewarded by the discriminator to not train on them. Finally, \citet{xu2019positive} reformulate discriminator training as Positive-Unlabeled (PU) learning which can be seen as a way of combating False Negatives. 

\section{Experiment setup}

\subsection{The BabyAI Environment}\label{subsec:babyai-tasks}

BabyAI is a deterministic, partially-observable 2D gridworld based on MiniGrid. The natural-looking language instruction is supplied as a string.  At each time step, the agent receives a visual input of the $7 \times 7$ cells in front of it.  Precise details of the visual input are described in \citep{babyai_iclr19}. We report performances on the following four single-room levels.

\textit{GoToRedBall} is the simplest level among the four and can be solved purely from visual inputs.  An agent is tasked with \texttt{go to the/a red ball} in the presence of other distracting objects of other colors and shapes.

\textit{GoToLocal} extends GoToRedBall. An agent is tasked with instructions of the form \texttt{go to the <color> <object>}, where \texttt{<color>} and \texttt{<object>} are no longer limited to \texttt{red} and \texttt{ball}. As a consequence, this level can no longer be solved by using just visual inputs -- a given instruction has to be parsed and understood.

\textit{PickupLoc} and \textit{PutNextLocal} are harder tasks which additionally require to equip the reward model with memory. In PickupLoc the instructions refer to objects not just by their type and color but also by their location relative to the initial position of the agent, e.g. \texttt{go to the red ball in front of you}. In the case of PutNextLocal, the instruction requires putting an object next to another object, each described with a type and color, e.g. \texttt{put the blue ball next to the blue key}. Since putting \texttt{object1} next to \texttt{object2} is not the same as putting \texttt{object2} next to \texttt{object1}, the reward model needs to remember the trajectory in order to give a correct reward at the end.

\subsection{Episode Termination}\label{sec:early_termination}

In the BabyAI platform the episode is terminated when the task is solved, or after a maximum allowed number of steps is reached, which is a standard practice for similar RL setups. However, in the GAIL setup, no environment reward is available. The naive approach would be to always run every episode for the maximum number of steps.

We let the agent perform a special \emph{Done} action. At the same time, all expert demonstrations are padded with this action as the last one (i.e. just before termination of the episode). It makes the agent perform the Done action when it considers the task to be done and it can be used to terminate the episode during training. It turns out that it significantly speeds up the training procedure as it collects richer data (more episodes for the same number of frames experienced by the agent). Done termination is used for all methods, including baselines.

The approach is related to actor early stopping introduced by \citet{zolna2019taskrelevant}. The difference is that we terminate episodes earlier based on the policy's predictions while the cited method terminates based on the discriminator's predictions.

\begin{table*}[ht]
\centering
\caption{Success rate for different algorithms. For each task we report performance for three expert demonstrations sets of different sizes (hence three columns are allocated for each task). The largest set is the minimum necessary demonstration set needed to solve the task using BC (the column headed with $1$). The other two are 8~times and 64~times smaller and headed with $\frac{1}{8}$ and $\frac{1}{64}$, respectively. A task is considered solved if the agent achieves more than 99\% success rate (bold values). In \emph{our method} we combine the Agent FC the Done Detector techniques.}
\smallskip
\small
\begin{tabular}{l||ccc|ccc|ccc}
 & \multicolumn{3}{c|}{\bf GoToLocal} & \multicolumn{3}{c|}{\bf PickupLoc} & \multicolumn{3}{c}{\bf PutNextLocal} \\
 {\bf Model} & $\frac{1}{64}$ & $\frac{1}{8}$ & $1$ & $\frac{1}{64}$ & $\frac{1}{8}$ & $1$ & $\frac{1}{64}$ & $\frac{1}{8}$ & $1$ \\
 \midrule \midrule
 Behavioral Cloning & 67.3 & 90.8 & \textbf{99.9} & 65.2 & 95.9 & \textbf{100.0} & 26.8 & 77.9 & \textbf{99.9} \\
 \midrule
 Baseline GAIL & 52.3 & 86.6 & 98.7 & 60.4 & 90.4 & 98.2 & 26.9 & 80.8 & 94.9 \\
 Oracle Filtering & \textbf{99.5} & \textbf{99.5} & \textbf{99.7} & 98.2 & \textbf{99.5} & \textbf{99.5} & 66.7 & 93.1 & 97.4 \\
  \midrule \midrule
 Our method & 98.9 & \textbf{99.7} & \textbf{99.8} & 96.5 & \textbf{99.2} & \textbf{99.3} & 98.2 & \textbf{99.6} & \textbf{99.3} \\
\end{tabular}
\label{tab:results}
\end{table*}

\subsection{Architecture and Training}

In our experiments, as mentioned in Section~\ref{subsec:conditioned}, the agent is provided with a non-zero reward only when it finishes an episode by performing a Done action described in the previous section. This makes rewards sparse and similar to the original rewards used in BabyAI platform to train RL algorithms. This choice allows us to use the agent architecture and the hyperparameters from the original BabyAI paper \citep{babyai_iclr19} without extra tuning.

\begin{figure}[t]
    \centering
    \begin{minipage}{0.45\linewidth}
    \centering
    \includegraphics[width=\linewidth]{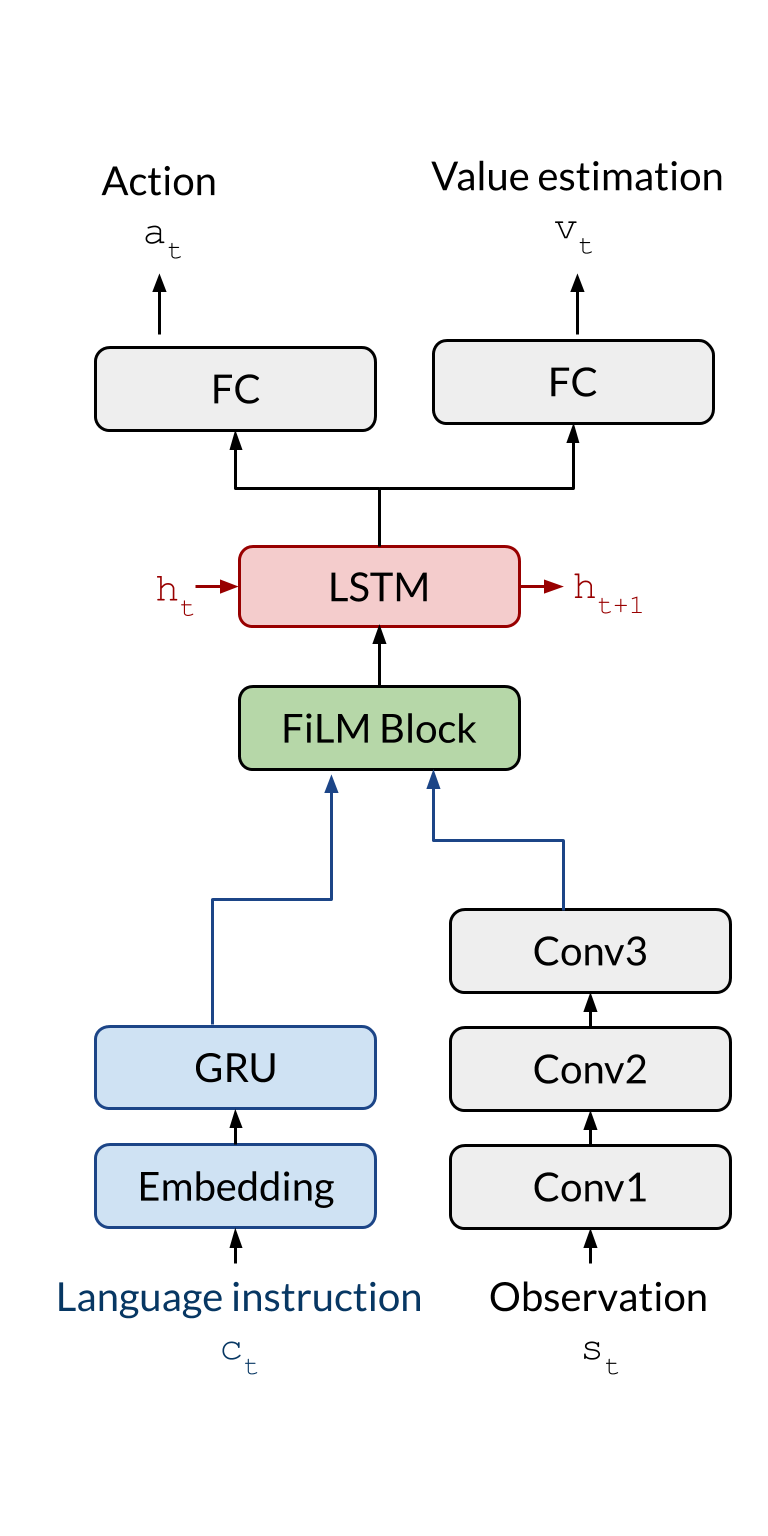}
    \small
    (a) Generator\\ \vspace{0.05cm}
    \label{fig:generator}
    \end{minipage}
    \hfill
    \begin{minipage}{0.45\linewidth}
    \centering
    \includegraphics[width=\linewidth]{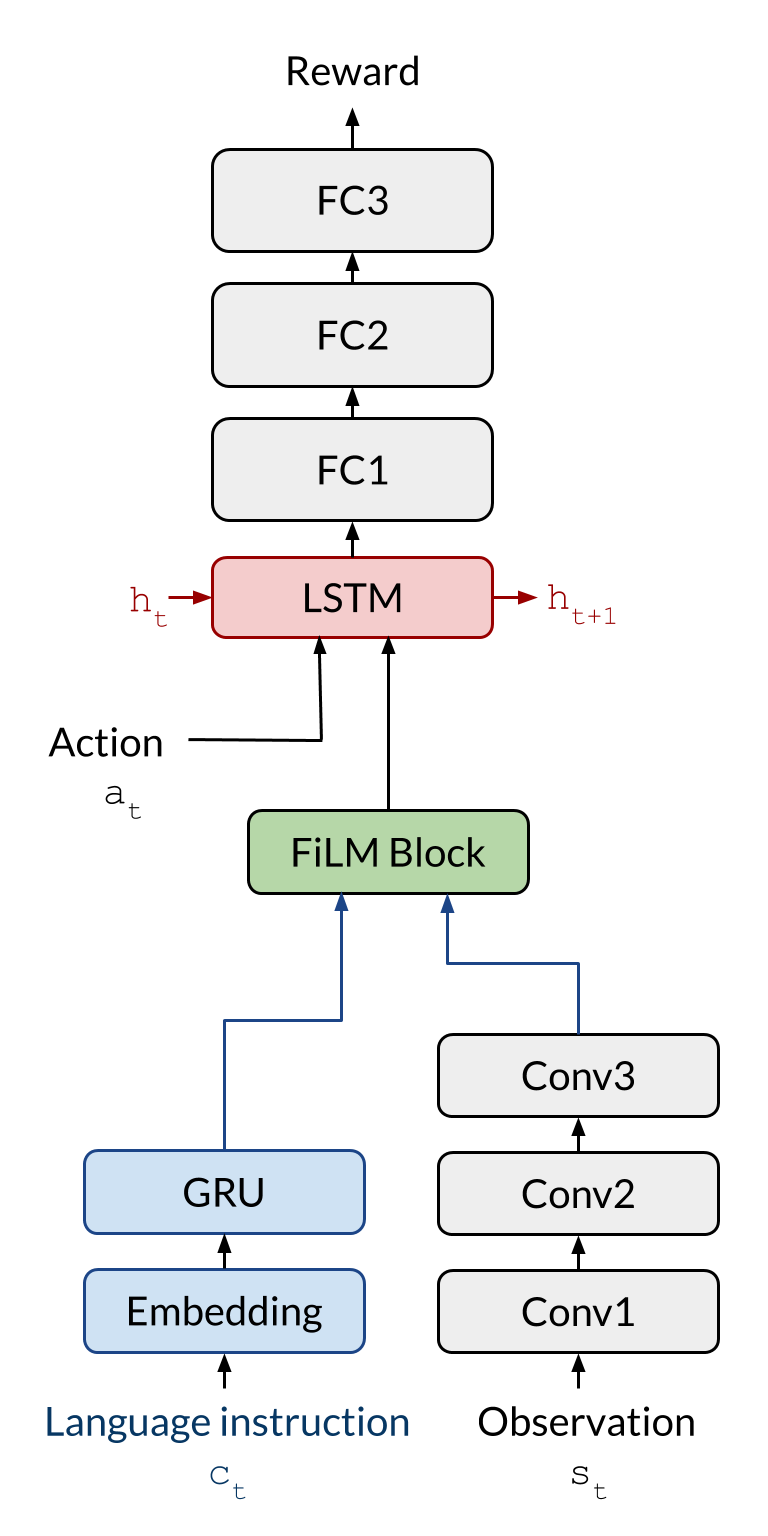}
    \small
    (b) Discriminator\\ \vspace{0.05cm}
    \label{fig:discriminator}
    \end{minipage}
    \caption{Model architectures. Best viewed in color.}
    \label{fig:arch}
\end{figure}

\paragraph{Actor-critic} The model underlying the RL agent is presented in Figure~\ref{fig:arch}(a). It consists of standard components to predict the next action based on the current observation, the memory of the past observations and the instruction. It uses GRU to encode the instruction and a convolutional network with two batch-normalized FiLM layers to jointly process the observation and the instruction. An LSTM memory is used to integrate representations produced by the FiLM module at each step. It uses a memory of 128 units and encodes the instruction with a unidirectional GRU.

\paragraph{Discriminator} The discriminator architecture (see Figure~\ref{fig:arch}(b)) is similar to the actor-critic model for symmetry and efficiency reasons. We just replace the final actor and critic layers by 3-layers MLP, placed after the FiLM block. The LSTM here takes as input not only the FiLM embedding but also a one-hot action vector (both concatenated). 

\paragraph{Training} In all our experiments, we use the same hyperparameters for actor training as used for RL training in \citep{babyai_iclr19}. For both models, we used the Adam optimizer with the hyperparameters $\alpha = 10^{-4}, \beta_1 = 0.9, \beta_2 = 0.999$ and $\epsilon = 10^{-5}$. We used the Proximal Policy Optimization algorithm with parallelized data collection: we performed 4 epochs of PPO using 64 rollouts of length 40 collected with multiple processes. We truncated the backpropagation through time at 20 steps for both actor-critic and discriminator.

During learning the agent is evaluated every 100 updates on 500 random episodes. The learning procedure is terminated after 10 successful evaluations (99\% success rate) or after 48 hours. We run each configuration considered in the paper with 3 different seeds, and report the average success rate. The variances between seeds are almost always below $0.2\%$ for solved tasks.

\subsection{Behavioral Cloning Performance}

We use BC results as the baseline for demonstration efficiency. First, for each task we tested how many demonstrations are needed to solve the task (achieve more than 99\% success rate) using BC. Our implementation of BC needs 125, 250, and 354 (in thousands) demonstrations for GoToLocal, PutNextLocal, and PickupLoc, respectively. These results are very close to the values reported in the original BabyAI paper \citet{babyai_iclr19}, which is expected as the architectures and hyperparameters are the same.

Since BC is the most basic IL method, the largest demonstrations sets considered in our experiments have the same number of demonstrations that was required to solve the tasks using BC. We also consider two subsets, one and two orders of magnitude smaller (precisely, 8 and 64 times smaller) for each task. These serve as a testbed for more efficient methods.

\subsection{Auxiliary Reward Mixing}

In Section~\ref{sec:false_negatives} four different losses that can be used to train the discriminators were introduced. Instead of training four separate models, we train just one neural network with multiple heads, each trained with one loss. We average all rewards used with equal weights to obtain the final reward used to train the agent.

\section{Results}

\subsection{Baseline GAIL and Oracle Filtering}

Our first contribution is to diagnose if false negatives impede adversarial imitation learning and how strong the effect is.

To do that, we first run a naive memory-equipped adaptation of GAIL, or Baseline GAIL, and demonstrate that it does not converge to the required performance level (see Table~\ref{tab:results}). Even when the full set of demonstrations is used, which is enough for BC to solve the task, Baseline GAIL does not reliably achieve $99\%$ success rate.

Then, we use the OF technique to diagnose the impact of the FN problem. The results are also presented in Table~\ref{tab:results}. OF clearly improves the performance and all levels are solved using an order of magnitude (8 times) fewer expert demonstrations than BC. For GoToLocal task even 64 times fewer demonstrations is enough. We reiterate that this significant improvement is achieved by only filtering out successful agent trajectories, which experimentally proves that the FN problem may be a major limiting factor for GAIL performance.

\subsection{Fake Conditioning}

Once the negative effect of the FN problem is confirmed, we test FC technique -- the solution proposed in Section~\ref{sec:false_negatives}.

In Table~\ref{tab:results}, the results for Agent FC along with Done Detector are referred as \emph{our method}. The method allows us to solve all tasks with an order of magnitude fewer demonstrations than BC. Even when 64 times smaller expert dataset is used, the obtained agent achieves over 96\% success rate for all tasks. The performance on PutNextLocal is better than that of the OF GAIL method despite the fact that the latter needs the environment rewards.

We experimentally found that simultaneous use of Agent FC along with Done Detector works the best for the considered task suite and we will refer to this particular combination as \emph{our method} in the rest of the paper. However, other combinations built with the use of FC methods also prove to be very effective and significantly outperform Baseline GAIL. The comparisons between them and detailed results are presented in Section~\ref{subsec:add-auxiliary-rewards}.

\subsection{FN Problem in Single Instruction Case}

We have so far focused on multi-goal tasks for which the FC technique is well suited. However, the FN problem is not specific to such tasks and can potentially hinder training when a single-goal (unconditioned) task is considered. In this section, we show that the FN problem is indeed a general problem and we propose a way to apply the FC technique in the single-goal case.

FC cannot be naively applied for unconditioned tasks because they have one fixed goal (instruction), and hence no fake instructions can be generated. However, for a given single-goal task, we can build a more complex multi-goal task where only one of the possible goals is the original task. Then, the FC technique can be applied to solve the thereby constructed task. Once a well-performing agent is obtained for the multi-goal task, it can also solve the original task as it is one among all goals the agent can be conditioned on.

GoToRedBall task is a simplified version of GoToLocal task where the agent is always given the same instruction (see Section~\ref{subsec:babyai-tasks}). We will test the idea described in the previous paragraph, i.e. we train our agent on the GoToLocal task using Agent FC method (with Done Detector) and test the trained agent on GoToRedBall.

Baseline GAIL and OF GAIL agents are trained using demonstration sets with only GoToRedBall instructions. For a fair comparison, the same number of demonstrations are used to train agents on GoToLocal and GoToRedBall. We note that even though the total numbers of demonstrations are the same, GoToLocal agent is trained with many fewer demonstrations with \texttt{go to red ball} as the instruction. The results are presented in Table~\ref{tab:gtrb}.

\begin{table}[h]
\caption{The results show success rate on GoToRedBall. We report performance for three expert demonstrations sets of different sizes (in three columns). The largest one is the minimum necessary demonstration set needed to solve GoToLocal tasks using BC. \emph{Our method} stands for GAIL enhanced with Agent FC and Done Detector and trained using GoToLocal demonstrations.}
\vspace{-0.1cm}
\smallskip
\small
\centering
\begin{tabular}{l||ccc}
 & \multicolumn{3}{c}{\bf GoToRedBall} \\
 {\bf Model} & $\frac{1}{64}$ & $\frac{1}{8}$ & $1$ \\
 \midrule \midrule
 Baseline GAIL & 78.5 & 97.4 & \textbf{99.4} \\
 OF GAIL & \textbf{99.7} & \textbf{99.8} & \textbf{99.8} \\
 \midrule
 Our method & 98.9 & \textbf{99.3} & \textbf{99.4} \\
\end{tabular}
\label{tab:gtrb}
\end{table}

OF GAIL clearly outperforms Baseline GAIL, which confirms our hypothesis that false negatives can have a negative impact on the training also in the single-goal case.

The agent trained with FC technique to solve GoToLocal achieves around 99\% when evaluated on the GoToRedBall which is similar to OF GAIL agent and significantly better than the Baseline GAIL agent. On top of that, the FC agent can also solve all the rest of GoToLocal instructions which the baseline methods can not. The performance difference on GoToRedBall between our method and Baseline GAIL is larger for smaller expert data sizes.

\section{Additional Studies}\label{sec:additional_studies}

\subsection{Sub-Trajectories}\label{subsec:exp-sub-trajectories}

As described in Section~\ref{subsec:conditioned}, the agent is rewarded only at the very end of an episode, i.e. once the full trajectory has been provided, and the discriminator is trained using full trajectories only. However, one can argue that a straightforward application of GAIL is to train the discriminator using incomplete sub-trajectories as well as complete ones. In that case, the discriminator loss is the following:
\begin{align}\label{eg:sub-trajectories}
    L^S_D(\theta) = & \mathbb{E}_{(c,\tau) \sim B^{sub}_{agent} \cup B_{agent}} - \log(1 - D_\theta(c,\tau)) \\ \nonumber
                + & \mathbb{E}_{(c,\tau) \sim B^{sub}_{expert} \cup B_{expert}} - \log(D_\theta(c,\tau)),
\end{align}
where $B^{sub}_{agent}$ and $B^{sub}_{expert}$ consist of incomplete sub-trajectories for the agent and expert, respectively. Note that the elements of $B^{sub}_{expert}$ are used here as \emph{positive} examples in contrast to how Done Detector is trained.

We conducted the experiments analysing the effect of using incomplete sub-trajectories. We trained Baseline GAIL discriminators in two ways, with sub-trajectories, as in Equation~\ref{eg:sub-trajectories}, and using only full trajectories, as in Equation~\ref{eq:memory}. The result are presented in Table~\ref{tab:resultsFE}.

\begin{table}[h!]
\caption{Success rate for Baseline GAIL using sub-trajectories or only full trajectories. For each task, we used the minimum necessary demonstration set needed to solve the tasks using BC.}
\vspace{-0.1cm}
\smallskip
\small
\centering
\begin{tabular}{l||c|c|c}
 {\bf Trajectories} & {\bf GoToLocal} & {\bf PickupLoc} & {\bf PutNextLocal} \\
 \midrule \midrule
 Done only & 98.7 & 98.2 & 94.9 \\
 All & 93.6 & 95.3 & 85.0 \\
\end{tabular}
\label{tab:resultsFE}
\end{table}

\noindent Even though both methods fail to solve the tasks (FC or any auxiliary rewards are \emph{not} used in these experiments), it is clear that using incomplete sub-trajectories has a deteriorative effect. We hypothesize that it is due to the fact that short trajectories in $B^{sub}_{expert}$ are hard to discriminate from unsuccessful agent trajectories. Hence, the elements of $B^{sub}_{expert}$ should be treated as negatives (as in Done Detector), not positives. 

\subsection{Done Termination}

As mentioned in Section~\ref{sec:early_termination}, we terminate the episode when the special Done action is performed by the agent. This technical detail turned out to be very useful and critical to achieve good performance. It also significantly speeds up training procedure, because richer data is collected -- more episodes for the same number of frames experienced by the agent. The results showing our method's performance with and without Done termination are presented in Table~\ref{tab:resultsDT}.

\begin{table}[ht]
\caption{The results show the drop in success rate when Done termination is not used. Order of magnitude (8 times) fewer demonstrations than needed for BC are used, and Agent FC and Done Detector is applied. Similar result are achieved for other demonstration sizes and methods.}
\vspace{-0.1cm}
\smallskip
\small
\centering
\begin{tabular}{l||c|c|c}
 {\bf Terminating} & {\bf GoToLocal} & {\bf PickupLoc} & {\bf PutNextLocal} \\
 \midrule \midrule
 Yes & \textbf{99.7} & \textbf{99.2} & \textbf{99.6} \\
 No & 63.8 & 47.3 & 15.9 \\
\end{tabular}
\label{tab:resultsDT}
\end{table}

\noindent When Done termination is not used, the success rate drops significantly for all tasks, and gets down to around $15\%$ for PutNextLocal task.

\begin{table*}[t]
\centering
\caption{Success rate for different models. For each task we report performance for three expert demonstrations sets of different sizes (hence three columns are allocated for each task). The largest one is the minimum necessary demonstration set needed to solve the tasks using BC. Two smaller subsets are tested (8~times and 64~times smaller). A task is considered solved if the agent achieves more than 99\% success rate (bold values).}
\smallskip
\small
\begin{tabular}{llcc||ccc|ccc|ccc}
 & & {\bf Done} & {\bf Blank} &  \multicolumn{3}{c|}{\bf GoToLocal} & \multicolumn{3}{c|}{\bf PickupLoc} & \multicolumn{3}{c}{\bf PutNextLocal} \\
 & {\bf Model} & {\bf Detector} & {\bf Conditioning} & $\frac{1}{64}$ & $\frac{1}{8}$ & $1$ & $\frac{1}{64}$ & $\frac{1}{8}$ & $1$ & $\frac{1}{64}$ & $\frac{1}{8}$ & $1$ \\
 \midrule \midrule
 (a) & Baseline GAIL & -- & -- & 52.3 & 86.6 & 98.7 & 60.4 & 90.4 & 98.2 & 26.9 & 80.8 & 94.9 \\
  \midrule \midrule
 (b) & Agent FC & -- & -- & 90.3 & \textbf{99.6} & \textbf{99.5} & 89.5 & \textbf{99.0} & \textbf{99.5} & 91.4 & \textbf{99.1} & \textbf{99.1} \\
 (c) & Agent FC & + & -- & 98.9 & \textbf{99.7} & \textbf{99.8} & 96.5 & \textbf{99.2} & \textbf{99.3} & 98.2 & \textbf{99.6} & \textbf{99.3} \\
 (d) & Agent FC & -- & + & 64.2 & \textbf{99.2} & \textbf{99.7} & 85.3 & \textbf{99.0} & \textbf{99.4} & 59.4 & 98.9 & \textbf{99.5} \\
 (e) & Agent FC & + & + & 94.6 & \textbf{99.5} & \textbf{99.7} & 96.8 & \textbf{99.5} & \textbf{99.4} & 81.5 & \textbf{99.4} & \textbf{99.5} \\
 \midrule
 (f) & Expert FC & -- & -- & 89.0 & \textbf{99.4} & \textbf{99.4} & 72.5 & 87.3 & 94.6 & 91.0 & 97.9 & 53.1 \\
 (g) & Expert FC & + & --  & \textbf{99.4}  & \textbf{99.7} & \textbf{99.7} & 93.1 & 95.8 & 96.7 & 96.9 & \textbf{99.0}  & \textbf{99.6} \\
 (h) & Expert FC & -- & + & 94.5 & \textbf{99.7} & \textbf{99.8} & 88.9 & 94.4 & 94.1 & 37.6 & 83.3 & 78.5 \\
 (i) & Expert FC & + & + & \textbf{99.3} & \textbf{99.6} & \textbf{99.7} & 93.2 & 97.4 & 97.3 & 95.6 & 94.4 & 98.1 \\
 \midrule
 (j) & Agent FC + Expert FC & -- & -- & 91.6 & \textbf{99.5} & \textbf{99.6} & 86.1 & 98.8 & \textbf{99.5} & 90.1 & 98.5 & \textbf{99.3} \\
 (k) & Agent FC + Expert FC & + & -- & 98.3  & \textbf{99.6} & \textbf{99.7} & 93.0 & \textbf{99.3} & \textbf{99.3} & 95.5 & \textbf{99.3} & \textbf{99.8} \\
 (l) & Agent FC + Expert FC & -- & + & 89.2 & \textbf{99.5} & \textbf{99.7} & 86.4 & \textbf{99.0} & \textbf{99.4} & 82.4 & 98.5 & \textbf{99.2} \\
 (m) & Agent FC + Expert FC & + & + & 95.1 & \textbf{99.7} & \textbf{99.6} & 93.7 & \textbf{99.2} & \textbf{99.4} & 88.6 & \textbf{99.5} & \textbf{99.5} \\
\end{tabular}
\label{tab:results_full}
\end{table*}

\subsection{Auxiliary Rewards}\label{subsec:add-auxiliary-rewards}

In this section we present results comparing different approaches using FC. Specifically, we consider three models: Agent FC, Expert FC and Agent FC + Expert FC. Each of these models can be run in 4 variants: without auxiliary rewards, with Done Detector or Blank Conditioning, or with both of them. The results are presented in Table~\ref{tab:results_full}. We will number specific rows alphabetically (from (a) to (m)) to make referring simple.

We will first focus on auxiliary rewards. Applying Done Detector always leads to better performing agent (please compare no auxiliary reward variants (b), (f) and (j) with their improved versions (c), (g) and (k), respectively). On the other hand, Blank Conditioning helps in some cases and hurts in others. As mentioned before, Blank Conditioning can suffer from the FN problem and we hypothesise this is the main reason why Blank Conditioning is worse than Done Detector. Done Detector is so beneficial that additional adding Blank Conditioning is never significantly better. Hence, in the rest of this subsection we will focus mainly on variants with Done Detector and without Blank Conditioning (as the best performing ones), i.e. (c), (g) and (k).

The best variant of Agent FC (c) tends to be better than Expert FC (g), especially when larger demonstration sets are used. The difference is more pronounced when variants without Done Detector are considered. Using both Agent FC and Expert FC at the same time does not seem to provide any further benefits.

All FC-based methods with Done Detector perform clearly better than Baseline GAIL (a). Among the tested methods, Agent FC with Done Detector auxiliary reward (c) achieves the highest performance in most cases. Agent FC without any auxiliary rewards also perform very well. On the other hand, Expert FC needs Done Detector to achieve good results. When Done Detector is not used, the discriminator, which is trained on expert trajectories only, seems to get explioted by the agent and the final performance is sometimes even worse than naive GAIL. However, Expert FC with Done Detector are the only variants that solve GoToLocal with 64 fewer demonstrations.

We would like to note that Expert FC and Done Detector discriminators does not depend on agent performance. Hence, they can be pretrained before the agent's learning procedure starts. It means that additional improvements known from supervised learning can be simply added to aforementioned pretraining. For example, validation early stopping can be very beneficial to prevent over-fitting to fixed and limited expert demonstrations when Expert FC discriminator is trained. We leave that for future work.

The take-away message from this section is that FC methods generally outperform Baseline GAIL which experimentally proves that addressing the problem of false negatives is important to achieve well performing agents. The choice of particular FC method presented in the paper is of the lesser importance, however using Agent FC with auxiliary reward based on Done Detector seems to be the best choice.

\section{Conclusions}

We show that the problem of false negatives can significantly hinder the performance of adversarial imitation learning. We contribute an extensive analysis of the phenomenon and a diagnostic tool, Oracle Filtering, to measure its impact. The tool is fully general and can be applied to any task.

We propose Fake Conditioning, a method to overcome the problem, and we show that it significantly improves over baselines for multi-goal tasks. We also presented a way to apply the method in the single-goal case.

Finally, we showed that auxiliary rewards obtained with extra discriminators can further improve the agent performance.

\section*{Acknowledgments}
We thank Louis Maestrati, Charles Guille-Escuret, Baptiste Goujaud, Anirudh Srinivasan, David Venuto, Junhao Wang, Christopher Beckham, Anne-Flore Baron for useful discussions. Konrad \.Zo\l{}na is supported by the National Science Center, Poland (2017/27/N/ST6/00828, 2018/28/T/ST6/00211). This research was mostly performed at Mila with funding by the Government of Quebec and CIFAR, and enabled by Compute Canada (www.computecanada.ca).

\bibliography{references}

\begin{thebibliography}{}

\bibitem[\protect\citeauthoryear{Abbeel and Ng}{2004a}]{abbeel}
Abbeel, P., and Ng, A.~Y.
\newblock 2004a.
\newblock Apprenticeship learning via inverse reinforcement learning.
\newblock In {\em In Proceedings of the Twenty-first International Conference
  on Machine Learning}.
\newblock ACM Press.

\bibitem[\protect\citeauthoryear{Abbeel and
  Ng}{2004b}]{abbeel2004apprenticeship}
Abbeel, P., and Ng, A.~Y.
\newblock 2004b.
\newblock Apprenticeship learning via inverse reinforcement learning.
\newblock In {\em Proceedings of the twenty-first international conference on
  Machine learning}, ~1.
\newblock ACM.

\bibitem[\protect\citeauthoryear{Bahdanau \bgroup et al\mbox.\egroup
  }{2019}]{bahdanau_learning_2019}
Bahdanau, D.; Hill, F.; Leike, J.; Hughes, E.; Hosseini, A.; Kohli, P.; and
  Grefenstette, E.
\newblock 2019.
\newblock Learning to {Understand} {Goal} {Specifications} by {Modelling}
  {Reward}.
\newblock In {\em International {Conference} on {Learning} {Representations},
  {ICLR} 2019}.

\bibitem[\protect\citeauthoryear{Chaplot \bgroup et al\mbox.\egroup
  }{2017}]{chaplot}
Chaplot, D.~S.; Sathyendra, K.~M.; Pasumarthi, R.~K.; Rajagopal, D.; and
  Salakhutdinov, R.
\newblock 2017.
\newblock Gated-attention architectures for task-oriented language grounding.
\newblock {\em CoRR} abs/1706.07230.

\bibitem[\protect\citeauthoryear{Chevalier-Boisvert \bgroup et al\mbox.\egroup
  }{2018}]{babyai_iclr19}
Chevalier-Boisvert, M.; Bahdanau, D.; Lahlou, S.; Willems, L.; Saharia, C.;
  Nguyen, T.~H.; and Bengio, Y.
\newblock 2018.
\newblock {BabyAI}: A platform to study the sample efficiency of grounded
  language learning.
\newblock In {\em ICLR}.

\bibitem[\protect\citeauthoryear{Duvallet, Kollar, and
  Stentz}{2013}]{duvallet2013imitation}
Duvallet, F.; Kollar, T.; and Stentz, A.
\newblock 2013.
\newblock Imitation learning for natural language direction following through
  unknown environments.
\newblock In {\em 2013 IEEE International Conference on Robotics and
  Automation},  1047--1053.
\newblock IEEE.

\bibitem[\protect\citeauthoryear{Fu \bgroup et al\mbox.\egroup
  }{2019}]{fu_language_2019}
Fu, J.; Korattikara, A.; Levine, S.; and Guadarrama, S.
\newblock 2019.
\newblock From {Language} to {Goals}: {Inverse} {Reinforcement} {Learning} for
  {Vision}-{Based} {Instruction} {Following}.
\newblock {\em arXiv:1902.07742 [cs, stat]}.
\newblock arXiv: 1902.07742.

\bibitem[\protect\citeauthoryear{Goodfellow \bgroup et al\mbox.\egroup
  }{2014}]{gan}
Goodfellow, I.; Pouget-Abadie, J.; Mirza, M.; Xu, B.; Warde-Farley, D.; Ozair,
  S.; Courville, A.; and Bengio, Y.
\newblock 2014.
\newblock Generative adversarial nets.
\newblock In Ghahramani, Z.; Welling, M.; Cortes, C.; Lawrence, N.~D.; and
  Weinberger, K.~Q., eds., {\em Advances in Neural Information Processing
  Systems 27}. Curran Associates, Inc.
\newblock  2672--2680.

\bibitem[\protect\citeauthoryear{Hermann \bgroup et al\mbox.\egroup
  }{2017}]{hermann}
Hermann, K.~M.; Hill, F.; Green, S.; Wang, F.; Faulkner, R.; Soyer, H.;
  Szepesvari, D.; Czarnecki, W.~M.; Jaderberg, M.; Teplyashin, D.; Wainwright,
  M.; Apps, C.; Hassabis, D.; and Blunsom, P.
\newblock 2017.
\newblock Grounded language learning in a simulated 3d world.
\newblock {\em CoRR} abs/1706.06551.

\bibitem[\protect\citeauthoryear{Ho and Ermon}{2016}]{ermon_gail}
Ho, J., and Ermon, S.
\newblock 2016.
\newblock Generative adversarial imitation learning.
\newblock In {\em NeurIPS}.

\bibitem[\protect\citeauthoryear{Kostrikov \bgroup et al\mbox.\egroup
  }{2018}]{kostrikov2018discriminator}
Kostrikov, I.; Agrawal, K.~K.; Dwibedi, D.; Levine, S.; and Tompson, J.
\newblock 2018.
\newblock Discriminator-actor-critic: Addressing sample inefficiency and reward
  bias in adversarial imitation learning.
\newblock {\em arXiv preprint arXiv:1809.02925}.

\bibitem[\protect\citeauthoryear{Luketina \bgroup et al\mbox.\egroup
  }{2019}]{overview_language}
Luketina, J.; Nardelli, N.; Farquhar, G.; Foerster, J.~N.; Andreas, J.;
  Grefenstette, E.; Whiteson, S.; and Rockt{\"{a}}schel, T.
\newblock 2019.
\newblock A survey of reinforcement learning informed by natural language.
\newblock {\em arXiv}.

\bibitem[\protect\citeauthoryear{Mei, Bansal, and
  Walter}{2016}]{mei_listen_2016}
Mei, H.; Bansal, M.; and Walter, M.~R.
\newblock 2016.
\newblock Listen, {Attend}, and {Walk}: {Neural} {Mapping} of {Navigational}
  {Instructions} to {Action} {Sequences}.
\newblock In {\em Proceedings of the 2016 {AAAI} {Conference} on {Artificial}
  {Intelligence}}.

\bibitem[\protect\citeauthoryear{Mirowski \bgroup et al\mbox.\egroup
  }{2018}]{street_bc}
Mirowski, P.; Grimes, M.~K.; Malinowski, M.; Hermann, K.~M.; Anderson, K.;
  Teplyashin, D.; Simonyan, K.; Kavukcuoglu, K.; Zisserman, A.; and Hadsell, R.
\newblock 2018.
\newblock Learning to navigate in cities without a map.
\newblock {\em CoRR} abs/1804.00168.

\bibitem[\protect\citeauthoryear{Ng and Russell}{2000}]{ng}
Ng, A.~Y., and Russell, S.~J.
\newblock 2000.
\newblock Algorithms for inverse reinforcement learning.
\newblock In {\em Proceedings of the Seventeenth International Conference on
  Machine Learning}, ICML '00,  663--670.
\newblock San Francisco, CA, USA: Morgan Kaufmann Publishers Inc.

\bibitem[\protect\citeauthoryear{Pomerleau}{1989}]{pomerleau89alvinn}
Pomerleau, D.~A.
\newblock 1989.
\newblock Alvinn: An autonomous land vehicle in a neural network.
\newblock In {\em NIPS}.

\bibitem[\protect\citeauthoryear{Ross, Gordon, and
  Bagnell}{2011}]{ross2011reduction}
Ross, S.; Gordon, G.; and Bagnell, D.
\newblock 2011.
\newblock A reduction of imitation learning and structured prediction to
  no-regret online learning.
\newblock In {\em Proceedings of the fourteenth international conference on
  artificial intelligence and statistics},  627--635.

\bibitem[\protect\citeauthoryear{Winograd}{1972}]{winograd}
Winograd, T.
\newblock 1972.
\newblock {\em Understanding Natural Language}.
\newblock Orlando, FL, USA: Academic Press, Inc.

\bibitem[\protect\citeauthoryear{Xu and Denil}{2019}]{xu2019positive}
Xu, D., and Denil, M.
\newblock 2019.
\newblock Positive-unlabeled reward learning.
\newblock {\em arXiv preprint arXiv:1911.00459}.

\bibitem[\protect\citeauthoryear{Ziebart \bgroup et al\mbox.\egroup
  }{2008}]{ziebart_entropy}
Ziebart, B.~D.; Maas, A.; Bagnell, J.~A.; and Dey, A.~K.
\newblock 2008.
\newblock Maximum entropy inverse reinforcement learning.
\newblock In {\em AAAI}.

\bibitem[\protect\citeauthoryear{Zolna \bgroup et al\mbox.\egroup
  }{2019}]{zolna2019taskrelevant}
Zolna, K.; Reed, S.; Novikov, A.; Colmenarej, S.~G.; Budden, D.; Cabi, S.;
  Denil, M.; de~Freitas, N.; and Wang, Z.
\newblock 2019.
\newblock Task-relevant adversarial imitation learning.

\end{thebibliography}
\bibliographystyle{aaai}

\end{document}